\documentclass{ametsocV6.1}
\usepackage{graphicx}
\usepackage{amsmath}
\usepackage{amssymb}
\usepackage{multirow}
\usepackage{natbib}
\usepackage{tabularx}
\usepackage{pdflscape}
\usepackage{makecell}
\usepackage{array}
\usepackage{caption}
\usepackage{circledtext}
\usepackage{color, colortbl} %
\usepackage{algorithm}
\usepackage{algpseudocode}
\usepackage{amsfonts}
\usepackage{makecell} 
\usepackage{fix-cm} % Allows arbitrary font sizes
\usepackage{tablefootnote}
\usepackage{lscape}
\title{Example-Based Concept Analysis Framework \\ for Deep Weather Forecast Models}

\authors{Soyeon Kim,\aff{a}
Junho Choi,\aff{a} 
Subeen Lee,\aff{a} 
and Jaesik Choi,\aff{a,b}
\correspondingauthor{Jaesik Choi, jaesik.choi@kaist.ac.kr}
}

\affiliation{\aff{a}{Kim Jaechul Graduate School of Artificial Intelligence, Korea Advanced Institute of Science and Technology, Seongnam, Gyeonggi, South Korea}\\
\aff{b}{INEEJI, Seongnam, Gyeonggi, South Korea}
}

\abstract{
To improve the trustworthiness of an AI model, finding consistent, understandable representations of its inference process is essential. This understanding is particularly important in high-stakes operations such as weather forecasting, where the identification of underlying meteorological mechanisms is as critical as the accuracy of the predictions. Despite the growing literature that addresses this issue through explainable AI, the applicability of their solutions is often limited due to their AI-centric development. To fill this gap, we follow a user-centric process to develop an example-based concept analysis framework, which identifies cases that follow a similar inference process as the target instance in a target model and presents them in a user-comprehensible format. Our framework provides the users with visually and conceptually analogous examples, including the probability of concept assignment to resolve ambiguities in weather mechanisms. To bridge the gap between vector representations identified from models and human-understandable explanations, we compile a human-annotated concept dataset and implement a user interface to assist domain experts involved in the the framework development.
}

\begin{document}

%% Necessary!
\maketitle

%%%%%%%%%%%%%%%%%%%%%%%%%%%%%%%%%%%%%%%%%%%%%%%%%%%%%%%%%%%%%%%%%%%%%
% SIGNIFICANCE STATEMENT/CAPSULE SUMMARY
%%%%%%%%%%%%%%%%%%%%%%%%%%%%%%%%%%%%%%%%%%%%%%%%%%%%%%%%%%%%%%%%%%%%%
%
% If you are including an optional significance statement for a journal article or a required capsule summary for BAMS 
% (see www.ametsoc.org/ams/index.cfm/publications/authors/journal-and-bams-authors/formatting-and-manuscript-components for details), 
% please apply the necessary command as shown below:
%
% Significance Statement (all journals except BAMS)
%
\statement
This study investigates deep neural networks' (DNNs) ability to encode semantic patterns of precipitation mechanisms and aims to provide a ready-to-deploy explainable artificial intelligence (XAI) tool. Key findings reveal that DNNs can extract nonlinear precipitation mechanisms and represent semantically meaningful meteorological attributes. The concept explanations align with expert perceptions, enhancing the interpretability and trustworthiness of model predictions. These findings demonstrate DNNs' potential to provide insightful, explainable predictions in meteorology, improving the trustworthiness of DNNs for practitioners. Follow-up research could involve refining the XAI framework, exploring its application for other meteorological phenomena, regions or scales, and integrating it with operational systems to assess the strengths and limitations in real-world scenarios.

%	 Enter significance statement here, no more than 120 words. See \url{www.ametsoc.org/index.cfm/ams/publications/author-information/significance-statements/} for details.
%
%% Capsule (BAMS only)
%%
% \capsule
% We build an explainable artificial intelligence (AI) framework grounded on the expert knowledge in a supervised scheme to provide human-understandable explanation to forecasters and enhance the trustworthiness of AI models.
%       Enter BAMS capsule here, no more than 30 words. See \url{www.ametsoc.org/index.cfm/ams/publications/author-information/formatting-and-manuscript-components/#capsule} for details.
%
%% * * If using twocol mode, you will need to use the commands "twocolsig" and "twocolcapsule" in place of "sig" and "capsule"
%%      to ensure that the text box correctly spans across both columns.
%

%%%%%%%%%%%%%%%%%%%%%%%%%%%%%%%%%%%%%%%%%%%%%%%%%%%%%%%%%%%%%%%%%%%%%
% MAIN BODY OF PAPER
%%%%%%%%%%%%%%%%%%%%%%%%%%%%%%%%%%%%%%%%%%%%%%%%%%%%%%%%%%%%%%%%%%%%%
%

\section{Introduction}
Recent applications of deep neural networks (DNNs) in meteorology demonstrate superior predictive performance and computation cost compared to traditional numerical weather prediction (NWP) models~\citep{bi2023accurate,lam2023learning,tang2022mtsmae}. However, actual adoption of DNNs in operational forecasting is slow due to their black-box nature: the high stakes associated with incorrect predictions require practitioners to have an intimate understanding of the inference process, an aspect that typical DNNs cannot address. A significant number of recent studies attempt to resolve this issue through explainable artificial intelligence (XAI)~\citep{yang2024interpretable, kim2023explainable, toms2020physically, mcgovern2019making, gagne2019interpretable}.

Existing applications of XAI in meteorology are often developed from the perspective of AI experts, reducing the utility of explanations for domain users. One solution to this problem is collaborating with the user population, which has been shown to offer significant benefits \citep{ravuri2021skilful}. This study builds upon this notion by constructing a user-centric XAI framework with an experts-in-the-loop approach, cooperating with experts at the Korea Meteorological Agency (KMA) and the National Institute of Meteorological Sciences (NIMS). Given that typical XAI methods are difficult for humans to comprehend~\citep{kim2023explainable}, we incorporate example-based explanation (explaining through samples that satisfy some criteria) and concept explanation (explaining through human-understandable semantics).
We also design a user interface to enhance the suitability of the framework for real applications, and perform case studies to measure the alignment between the generated explanations and domain knowledge.

For the explanation task, we tackle the question whether DNNs’ representations encode semantically meaningful nonlinear precipitation mechanisms, a task that is yet to be addressed in prior literature \citep{kurihana2024identifying,jo2020classification,park2021diverse}.
Specifically, we address the following two topics:
Can we detect nonlinear precipitation mechanisms from trained DNNs?
Can we identify the presence of meaningful meteorological attributes such as convectional, frontal, orographic, and convergence mechanisms from internal representation space in trained DNNs?

The rest of this paper is organized as follows.
Section \ref{related} provides an overview of past literature on example-based, concept-based, and user-centric explanations.
Section \ref{method} outlines the example-based concept analysis framework, human annotation process for meteorological data, and user interface design.
Section \ref{experiment} discusses the experimental setup, including model and data.
Section \ref{results} assesses the results of the experiments both quantitatively and qualitatively before discussing the implications of the findings.
Section \ref{conclusion} concludes the paper.

\section{Related Work}\label{related}

\subsection{Example-Based Explanation}
Example-based explanation is a popular XAI method that is easy for layman users to understand, a characteristic essential to user-centric XAI~\citep{molnar2020interpretable}. Nearest neighbor search is one such method, but the results can vary by the proximity metric \citep{johnson2016perceptual}. Euclidean distance in the feature space of DNN is more human perceptually close than sophisticated similarity measures in the input level~\citep{zhang2018unreasonable, amir2021understanding}. A recent study demonstrates that configuration distance - the Hamming distance between activation status of feature vectors - is also semantically aligned with human perception~\citep{chang2024understanding}. However, since all metrics inevitably require computing pairwise distances, nearest neighbor search does not scale to high-dimensional data. To address this issue, we implement a nearest neighbor search with dimensionality reduction.

\subsection{Concept Explanation}
A concept refers to semantic representations such as objects, shapes, textures, and colors~\citep {schwalbe2022concept}. Concept analysis is applied in numerous domains since it is intrinsically human-intelligible~\citep{schut2023bridging, cai2019human}. Another advantage of concept explanation is the targeted concepts does not need to be intrinsic to the target model's task: most studies use human annotations to assign meaningful concepts~\citep{kim2018interpretability}, which may not necessarily be directly associated with class labels. Therefore, we can probe the high-level semantic information from the internal representations of the AI models. In meteorology, concept analysis has been studied to identify weather mechanisms captured in AI models, such as the eye of the typhoon in a DNN~\citep {sprague2019interpretable}. However, to the best of our knowledge, representations captured in spatiotemporal models has not yet been studied in previous papers. Our work builds upon \textit{Testing with Concept Activation Vectors} (TCAV)~\citep{kim2018interpretability} to identify spatiotemporal patterns for rainfall mechanisms.

\subsection{Concept Prober}
Probing is a technique to understand the concepts captured in trained models~\citep{alain2016understanding} and the influence of these representations on model prediction~\citep{belinkov2022probing}. Approaches include using (a) the probability from independently trained support vector machine (SVM) classifiers for each concept \citep{kim2018interpretability}, (b) mutual information between representations and target labels \citep{pimentel2020information}, and (c) Gaussian processes \citep{wang2024gaussian}.
Probing is applied in tasks such as finding factual associations of large language models~\citep{meng2022locating}, analyzing causality perspective~\citep{vig2020investigating}, validating model hallucination~\citep{azaria2023internal}, or identifying linguistic structures~\citep{hennigen2020intrinsic}. \cite{joung2024probing} applies probing classifiers to study counterfactuals in the image domain.
To the best of our knowledge, none of these methods have been adapted to weather systems, especially on object-instance units of a trained model whose inputs and outputs are separated in time. We fill this gap by applying concept probing to identify rainfall mechanisms.

\section{Method}\label{method}
The proposed framework consists of a probabilistic concept prober (Section~\ref{method}.\ref{method:framework}.\ref{method:prober}) and neighbor search engine (Section~\ref{method}.\ref{method:framework}.\ref{method:nn}). We evaluate the identified concepts by adapting the existing methods to segmentation model architectures (Section~\ref{method}.\ref{method:metric}) with a human annotated concept dataset (Section~\ref{method}.\ref{method:label}).

\subsection{Example-Based Concept Explanation Framework}\label{method:framework}

\subsubsection{Probabilistic Concept Prober.}\label{method:prober}
\begin{figure}[h]
    \centering
    \includegraphics[width=\linewidth]{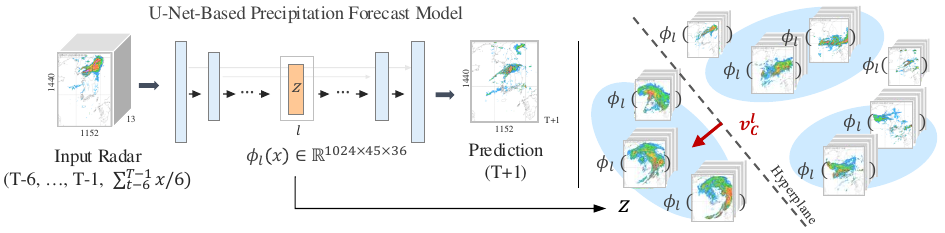}
    \caption{Illustration of concept prober training process.}
    \label{fig:tcav}
\end{figure}

Following the research of~\cite{kim2018interpretability}, we perform supervised concept analysis by training one-vs-all binary SVM classifiers on domain knowledge-grounded concept labels (Fig.~\ref{fig:tcav}). These classifiers are used for computing the probability of the presence of a particular concept in the feature vectors extracted from the target model's bottleneck layer.

The process of training individual concept probers is illustrated in Fig.~\ref{fig:tcav}.
First, we extract the feature vectors of the bottleneck layer of the target model (specifically, the output of the \textit{DownSample} layer described in Table~\ref{tab:model}) to use as the training data for the concept probers. 
Second, the concept probers are trained using the resized segmented activation vectors and the concept labels (Section~\ref{method}.\ref{method:label}). Each concept prober is trained in a one-vs-all manner using binarized concept labels. It should be noted that the same training data is used for the nearest neighbor search engine (Section~\ref{method}.\ref{method:framework}.\ref{method:nn})

\subsubsection{Nearest-Neighbor Search Engine.}\label{method:nn}
We build a nearest-neighbor search module based on Euclidean distance in the bottleneck feature space of the target model. Since the target model operates on high-resolution data, the dimension of the bottleneck layer is $1,024\times45\times36\approx1.65\times10^6$, leading to significant costs when computing distances across all pairs of data points. To address this issue, we reduce the features to semantically meaningful principal neuron components (PC) using the concept probers and relaxed decision regions (RDR)~\citep{chang2024understanding}. The RDR approach selects PC, which are relatively discriminatively activated with respect to the negative vector (average activation state of negative training samples). The activation of PC indicates the presence of semantic patterns in an instance. This alignment allows retaining significant semantic features with significantly reduced dimensions. 
We apply RDR for each prober to find PC for each concept, which is used in the actual nearest neighbor search algorithm shown in Algorithm~\ref{algo}. For a given query, we compute the logit probability of each concept using the probers and select the top $k_1$ concepts. We use the union of the concepts' PC, whose dimensionality $d$ is much smaller than dimensionality $D$ of the original space, to compute the distances. Given the sample size $N$ and concept size $C$,
the computation cost of pairwise Euclidean distance between dataset and one query sample in the original feature space is $\mathcal{O}(ND)$; that in the reduced space is $\mathcal{O}(Nd)$; and the computation for $C$ probers on the $n'$ nearest neighbors is $\mathcal{O}(n'CD)$. Thus, our method costs $\mathcal{O}(Nd + n'CD)\rightarrow\mathcal{O}(Nd)$ since $n'\ll N$, which is much smaller than $\mathcal{O}(ND + n'CD)\rightarrow\mathcal{O}(ND)$ since $d \ll D$. It demonstrates that the computational bottleneck is caused in nearest-neighbor search instead of the concept probing process, and that the dimensionality reduction approach alleviates the time complexity.
We observe in the experiments that using PC for dimensionality reduction causes only minor performance degradation compared to conventional algorithms such as empirical orthogonal functions(EOF) and principal component analysis(PCA) (Table~\ref{tab:runtime}). We hypothesize that this difference is caused by the underlying assumption of conventional methods, which find orthogonal vectors based on the magnitude of the variance in the covariance matrix of neuron activation. However, physical systems are not necessarily orthogonal~\citep{hannachi2007empirical}, which reduces their effectiveness compared to RDR. This comparison indicates that it worthwhile to reduce dimensions in the proposed fashion. Considering the trade-off between time complexity and and performance retention discussed in Table~\ref{tab:dimensionality} in Appendix~\ref{appx:dimensionality}, we set the hyperparameter for the number of PC as 300.

\begin{algorithm}[H]
\caption{Neighbor Search Engine using Principal Neuron Components}\label{algo}
\begin{algorithmic}[1]
\Require A dataset \(\textbf{X}\), a prober function \(f_c\) for concept \(c\), a dictionary \(R\) where key is a concept and value is principal neuron indices, a query sample \(\textbf{x}_q\), hyperparameter \(k_1\) for k concepts to consider, hyperparameter \(k_2\) for top-k nearest neighbors.
\Ensure Nearest neighbor sample indices \(S\).
\State \( Y_C \gets f_C(x_q) \) 
\Comment{Compute logit probabilities for all concepts \(C\).}
\State \( I \gets \text{argsort}(Y_C)[-k_1:] \) 
\State \( P \gets \bigcup_{i \in I} R[i] \) \Comment{Retrieve a list of principal neuron indices.}
\State \( \text{Dist} \gets ||\textbf{X}[P] - \textbf{x}_q[P]||^2 \)
\State \( S \gets \text{argsort}(\text{Dist})[:k_2] \) \Comment{Select \(k_2\) nearest neighbor samples.}
\State \Return S
\end{algorithmic}
\end{algorithm}

\begin{table}[H]
\setlength{\tabcolsep}{5pt}
\caption{Runtime comparison of principal neuron component-based neighbor search engine (PC-NSE).}
\centering
\begin{tabular}{llcccc}
\hline\hline
Embedding & \#Dim  & Runtime(sec) & Precision@3  & Precision@5 & Precision@10 \\
\hline
$Z$ 
    & 1024$\times$36$\times$45 
    & 6.80 
    & 0.471 $\pm$ 0.144 & 0.349 $\pm$ 0.114 & 0.231 $\pm$ 0.092 \\
$Z_{PCA}$
    & 300
    & 2.60
    & 0.391 $\pm$ 0.075 & 0.256 $\pm$ 0.066 & 0.143 $\pm$ 0.045 \\
$Z_{PC-NSE}$
    & 300 
    & \textbf{1.50}
    & 0.467 $\pm$ 0.180 & 0.317 $\pm$ 0.132 & 0.194 $\pm$ 0.092 \\
\hline
\end{tabular}
\label{tab:runtime}
    \vspace{0.1in}
    {\raggedright 
    % Runtime is measured per a single query sample, comparing relative performance of Precision@k which refers to the ratio of correct results among the top k nearest neighbor results using human annotation labels (see Section~\ref{method}.\ref{method:label}):
    % $ Precision @ k = \frac{|\text{correct samples among } k \text{ results}|}{k} $.
    % The experiment was conducted on an Intel Xeon Gold 6342 CPU @ 2.8GHz with 96 logical cores. The random seed is fixed at 42.} 
    Runtime is measured per a single query sample, comparing relative performance of Precision@k with top k nearest neighbor labels.
    The experiment is conducted on an Intel Xeon Gold 6342 CPU @ 2.8GHz with 96 logical cores. The random seed is fixed at 42.} 
    \par
\end{table}

\subsection{Human Annotation Concept Labels}\label{method:label}

% We create a weather mechanism label dataset based on several existing materials, including case studies based on four documents and the results of two related projects on model-based precipitation categorization (the first and the last four rows in Table~\ref{tab:reference}, respectively); daily post-hoc forecast analysis reports provided by KMA; and heavy rainfall classification reports provided by NIMS.
% In the case of the last document, daily post-hoc forecast analysis reports are described separately by a single rainfall system unit; instead, others are written by date-time unit. Thus, we annotate the object instance level in the daily reports and the date-time level in the others.
% The object instance level corresponds to the output feature of $\mathcal{W}\circ\phi_l$, while the date-time level corresponds to the output feature of $\phi_l$ in Fig.~\ref{fig:tcav}.}

We create a weather mechanism label dataset based on several materials, including daily post-hoc forecast analysis reports provided by KMA and heavy rainfall classification reports provided by NIMS. It should be noted that daily post-hoc forecast analysis reports are described by rainfall system units, which differs from the date-time unit used in other sources. We compile both information in the final dataset.

\subsection{Concept Evaluation for Segmentation Models}\label{method:metric}

% The importance score of a concept~\citep{kim2018interpretability, ghorbani2019towards} is the magnitude of change in a model's output caused by a shift in the direction of the corresponding concept activation vector (CAV)} in the feature space. Since the original method is designed for classifiers, we modify the metric for segmentation models by measuring the aggregated changes across the output. Inspired by \cite{kokhlikyan2020captum}}, we introduce a wrapper function $\Psi_k$ in Eq.~\eqref{eq:wrapper} that aggregates the classification result for class $k$ across the entire output of a segmentation model. The importance score in Eq.~\eqref{eq:importance_score} is computed} based on the sensitivity of the aggregate to small perturbations in the direction of the CAV in Eq.~\eqref{eq:sensitivity}. 

The importance score of a concept~\citep{kim2018interpretability, ghorbani2019towards} is the magnitude of change in a model's output caused by a shift in the direction of the corresponding concept activation vector (CAV) in the feature space. Since the original metric is designed for single-label classifiers, we modify it for segmentation models by measuring the aggregated changes across all outputs. Inspired by \cite{kokhlikyan2020captum}, we introduce a wrapper function $\Psi_k$ in Eq.~\eqref{eq:wrapper} that aggregates the result for class $k$ across the entire segmentation output. The importance score in Eq.~\eqref{eq:importance_score} is computed based on the sensitivity of the aggregate to small perturbations in the direction of the CAV in Eq.~\eqref{eq:sensitivity}. 

%We discuss additional variants of wrapper functions in Appendix~\ref{appx:wrapper}.

\begin{linenomath*}
\begin{equation}
\begin{aligned}
S_{c, k}(x) & =\lim _{\epsilon \rightarrow 0} \frac{\Psi_k\left(h_k\left(\phi(x)+\epsilon v_c\right)\right)-\Psi_k\left(h_k(\phi(x))\right)}{\epsilon} \\
% & =\lim _{\epsilon \rightarrow 0} \frac{\Psi_k\left(h_k\left(\phi(x)+\epsilon v_c\right)\right)-\Psi_k\left(h_k(\phi(x))\right)}{\epsilon v_c} v_c \\
& =\nabla \Psi_k\left(h_k(\phi(x))\right) \cdot v_c
\label{eq:sensitivity}
\end{aligned}
\end{equation}
\end{linenomath*}
\begin{linenomath*}
\begin{equation}
I_{c, k}=\frac{\left\| x \in X_k: S_{c, k}(x)>0 \right\|}{\left\| X_k \right\|}
\label{eq:importance_score}
\end{equation}
\end{linenomath*}
\begin{linenomath*}
\begin{equation}
\Psi_k\left(h, \phi, x\right)=\sum\limits_{i}^{W} \sum\limits_j^H\left(\phi \circ h_k\right)\left(x_{i, j}\right) \text {, s.t. } \underset{\mathrm{k} \in \mathrm{K}}{\arg \max }\left(\phi \circ h_k\right)\left(x_{i, j}\right)=k
\label{eq:wrapper}
\end{equation}
\end{linenomath*}

\noindent Given model $F$, input $x$, concept $c$, and class $k$, $\phi(x)$ is the composition function up to target layer $l$ and $h(z)$ is the composition function downstream from layer $l$, i.e., $F(x)=(\phi \circ h)(x)$. $h_k(z)$ is the function $h(z)$ with respect to target class $k$. 
$v_c$ is the CAV corresponding to $c$. 
$S_{c,k}(x)$ is the sensitivity of the model output to perturbation on $\phi(x)$ in the direction of $v_c$. 
$X_k$ is the set of samples whose ground truth includes $k$. 
$I_{c,k}$ is the importance score, which is the ratio between the number of samples with positive $S_{c,k}$ and the number of samples in $X_k$.
Note that the score range is dependent on the logit or loss values. In this study, our target model has a logit value range of $[0,\infty )$ and a loss range of $[0,1]$.

\section{Experiments}\label{experiment}

\subsection{Target Model and Data}
\paragraph{\textbf{}Model.}
The experiments are performed using an unpublished variant of DeepRaNE~\citep{ko2022effective} trained on 10-minute interval composited radar hybrid surface rainfall (HSR) data in Korea between 2018 and 2021. This model classifies precipitation into eight categories: 0-0.1, 0.1-1, 1-5, 5-10, 10-20, 20-25, 25-30, and 30 mm hr -1. 
Further details are provided in Appendix~\ref{appx:data}.

\paragraph{Data.}
The data and relevant parameters are provided by KMA. The training data of the segmentation model consists of processed hybrid surface radar (HSR) data and spatiotemporal information. The raw HSR data is provided in dBZ. Each raw HSR image is first scaled by dividing by 100, and is converted to radar reflectivity $Z$ using $Z=10^{dBZ\times0.1}$. $Z$ is then converted to rain rate $R$ using the Z-R relationship of $R=(Z/a)^{1/b}$ with parameters of $a=148, b=1.59$. Each instance of model input concatenates seven processed images at 10-minute intervals, from 60 minutes prior to the reference time(T), plus the 1-hour cumulative average. The input also includes latitude, longitude, and date information (month, day, and hour). 

Classification targets are based on lagged features with intervals of 60 minutes, conditioned on lead times of 1 to 6 hours. The ground truth is derived from averaging the previous 60 minutes of radar data.
For computational efficiency, the input images are downsampled from $2,304\times2,880$ to $1,152\times1,440$ pixels using max pooling, as it better preserves strong precipitation patterns compared to average pooling.
Normalization techniques vary by data type. Time information is min-max normalized to the range [0,1]. Latitude and longitude values are scaled to ranges (0.6911, 1] and (0.8899, 1], respectively. Radar rainfall intensity is normalized using a modified hyperbolic tangent function of $X_t = 0.5 \times \operatorname{Tanh}(0.01\times \frac{X_t-\mu_X}{\sigma_X})$ with $\mu_X=-0.01$ and $\sigma_Xs=4$, and is then scaled to the range (-0.8182, 1]. This approach is preferred for its robustness against outliers and faster convergence compared to traditional Z-score normalization.

\paragraph{Data Preprocessing.}
The bottleneck layer of our target model is sparsely activated, with only 280 out of 1,024 channels activating at least once across the entire training dataset. For the computational efficiency, we focus on these 280 channels
    \footnote{Channel pruning is used in various studies since activation sparsity is desirable for memory efficiency \citep{kurtz2020inducing}. For example, \cite{rhu2018compressing} introduce a zero-valued compression approach to leverage sparsity. In a probing-related study, \cite{hennigen2020intrinsic} reports that, on average, only a small proportion of encoded neurons are allocated to semantic features. However, \cite{gao2018dynamic} warns that channel pruning must be performed carefully, as the contributions of pruned channels are permanently lost. In our case, it is fairly reasonable that the eliminated channels would remain inactive given their lack of activation in the training dataset.}
. In addition to the data processing above, we recognize that individual precipitation systems at a single date and time should be treated independently. To address this issue, we further preprocess the data by separating the precipitation areas within a single input applying a segmentation algorithm (e.g., Watershed \citep{beucher1979use, beucher1992watershed, neubert2014compact}). The resulting segments are resized to a predefined size of $(C,H,W) = (280,9,9)$. This approach facilitates the identification of distinct rainfall mechanism-aware concepts and addresses the issue of high dependency on spatial information in pattern recognition.

\subsection{Experimental Settings}

\paragraph{Concept Probers.}
For each prober, we split the training and validation sets in a 9:1 ratio using a random seed of 42. Since the class labels are highly imbalanced, we perform stratified sampling to create an one-to-all binary classification dataset, with the positive and negative sets sampled from in and out-of-class data. 

We use SVM classifiers as concept probers and train them using stochastic gradient descent with logistic loss. To alleviate the high dimensionality problem, we use $L_{1}$ regularization for sparsity and efficient probing inference. We also calibrate the trained probers with Platt's Sigmoid method~\citep{platt1999probabilistic} and ensemble the output probabilities using five-fold cross validation (CV) on the test dataset to address the potential overconfidence issue. The final output is the averaged prediction probabilities of all CV pairs. The CAVs are extracted from the averaged coefficients of the ensemble models. We utilize \texttt{SGDclassifier} and \texttt{CalibratedClassifierCV} provided by \texttt{sklearn} python APIs~\citep{scikit-learn} for training.

\paragraph{Benchmarks.}\label{appx:gpp}
We compare our probers against \cite{joung2024probing} and \cite{wang2024gaussian}. 
\cite{joung2024probing} uses a simple multi-layer perceptron (MLP)-based nonlinear classifier with two fully-connected layers and rectified nonlinear activation function (ReLU) defined as: 
\begin{linenomath*}
\begin{equation}
    y_{c,i} \sim \operatorname{softmax}(W_c^{2nd}\cdot \operatorname{ReLU}(W_c^{1st} \cdot \phi(x_i)+b_c^{1st})+b_c^{2nd}).
    \label{eq:MLP}
\end{equation}
\end{linenomath*}
where $W_c$ and $b_c$ denote weight and bias of $c$-th concept prober.
To build the MLP prober, we use input size of 22,680, hidden size of 1,000, ReLU activation function, and Adam optimizer with learning rate of 0.001. Additionally, we add temperature scaler \citep{guo2017calibration} to calibrate the logits before the Sigmoid function.
For the Gaussian process-based prober (GPP), we use the default settings from the open-source code provided at github repository
    \footnote{\url{https://github.com/google-research/gpax/}}
. Due to the computational limitations of GPP, we reduce the data dimensionality from 22,680 to 100 using incremental PCA. The significant dimensionality reduction may contribute to the low performance of GPP. The predictive probability threshold is set to 0.5.

\section{Results and Discussion}
\label{results}

\subsection{Evaluation of Nearest Search Engine}\label{results:nn}

We set our algorithm to provide three nearest-neighbor samples ($k_1 = 3$), as well as the top 5 concepts and their probing predictive probabilities for each neighboring samples and the query instance. These hyperparameters are selected based on user interviews. However, the hyperparameter $k_1$ can be adjusted according to the user's preferences since the number of maximum nearest neighbors is not directly related to the algorithm's performance.

\begin{figure}[H]
    \centering    
    \includegraphics[width=\linewidth]{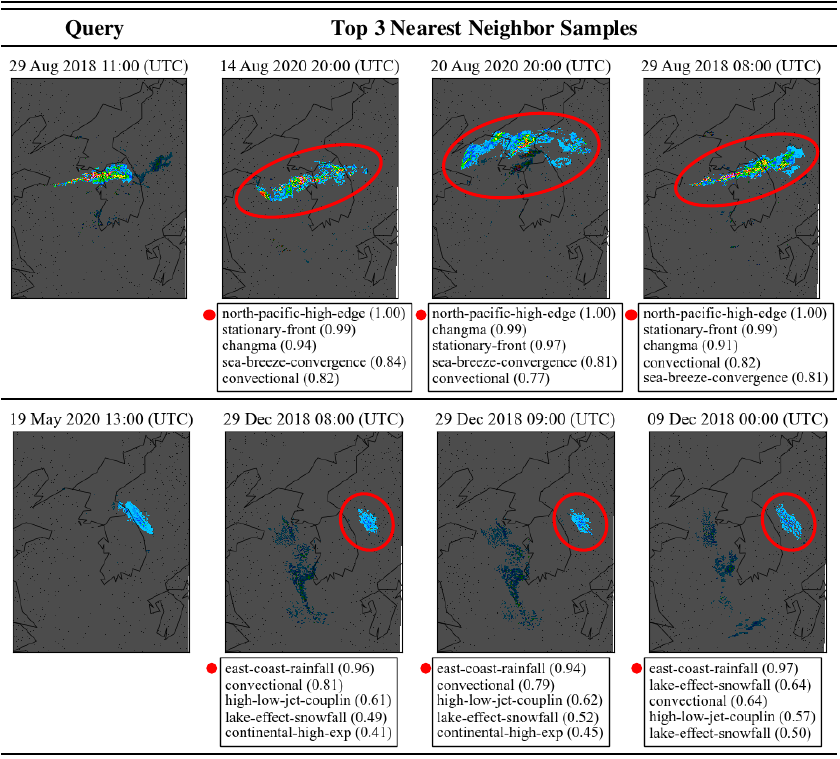}
    \caption{Examples of neighbor search engine and probabilistic concept explanations. Each row is a single query. The first column of each row are the query samples. The remaining columns are the three nearest neighbors of the queries. Each instance is reported with top-5 rainfall mechanism concepts in terms of prober's probability. The numerical values after each concept are the predictive probabilities from the corresponding prober.}
    \label{fig:concept}
\end{figure}

Fig.~\ref{fig:concept} presents one query example for heavy rainfall in summer and another for light rainfall in spring. 
The summer rainfall case, characterized by elongated east-west heavy rainfall, corresponds to the mechanisms of \texttt{stationary frontal heavy rainfall} and \texttt{the edge of the North Pacific High}. 
The selected neighbors also exhibit linear heavy rainfall patterns and are classified by the probe to represent concepts related to \texttt{stationary fronts} and \texttt{the edge of the North Pacific High} mechanisms.
\texttt{The edge of the North Pacific High} and \texttt{stationary frontal rainfall} are some of the main rainfall patterns in summer on the Korean Peninsula as the expansion and contraction of the North Pacific High control the location and the intensity of stationary frontal rainfall. 
For the second example, the pattern is classified as \texttt{east coast rainfall}, which occurs due to land-sea friction caused by easterly winds. The engine identifies similar cases representing various states of a similar mechanism, indicating that both the nearest neighbor search engine and the concept probers have been well-trained to fulfill their purposes.
Since our engine can identify samples with semantically similar mechanisms and provide probablistic interpretation of the mechanisms even if their visual form and precipitation intensity do not match exactly, the results may assist in analyzing heavy rainfall patterns.

\subsection{Evaluation of Concept Prober}\label{results:prober}
In Table~\ref{tab:prober}, we report the macro F1 score (Eq.~\ref{eq:macrof1}), the arithmetic average of F1 scores across classes (C), and the accuracy averaged over all classes with respect to the given test data and labels.

\begin{equation}
    \operatorname{Macro F1} = \cfrac{1}{C}\sum^C_{c=1} \operatorname{F1}_c
\label{eq:macrof1}
\end{equation}
\begin{equation}
    \operatorname{F1}_c = 2 \cdot \cfrac{\operatorname{Precision}_c \cdot \operatorname{Recall}_c}{\operatorname{Precision}_c +\operatorname{Recall}_c}
\end{equation}

We compare our probers with recently proposed probing architectures. As shown in Table~\ref{tab:prober}, the results demonstrate that SVM classifiers outperform other methods, making them well-suited for concept probing.

Although deeper architectures could potentially be used for probes, many previous studies employ simple linear classifiers or shallow multi-layer perceptrons (MLPs) (as described in Eq.~\ref{eq:MLP}) \citep{joung2024probing, hall2020tale, liu2019linguistic, hupkes2018visualisation, alain2016understanding}. This design choice is explained by the goal of a prober: verifying the effective encoding of target concepts in the feature space~\citep{belinkov2022probing}. If a prober becomes complex, it becomes difficult to determine whether the observed results stem from the intermediate layer representations of the model or from patterns additionally learned by the prober~\citep{hewitt2019structural, hupkes2018visualisation}. If a simple model cannot properly identify the presence of a concept based on a probablistically distributed feature space, it indicates that the concept is not captured by this space. 
Furthermore, if the model is well-trained, its intermediate feature space should approximate a Hilbert kernel space, making a simple linear classifier sufficient to identify the presence of specific conceptual properties. Our results support these notions.

\renewcommand{\arraystretch}{1}
\begin{table}[H]
\caption{Performance of concept prober.}\label{tab:prober}
\centering
\begin{tabular}{lll}
\hline\hline
Model &  Macro F1 & Accuracy\\
\hline
SVM~\citep{kim2018interpretability} & 0.7636 & 0.7610\\
2-Layer MLP~\citep{joung2024probing} & 0.5751 & 0.6925\\
GPP~\citep{wang2024gaussian} & 0.5693 & 0.5566 \\
% ResNet18~\citep{he2016deep}} & 000} & 000} \\
\textbf{SVM (calibrated ensemble}) &  \textbf{0.7700} & \textbf{0.7686}\\
\hline
\end{tabular}
\end{table}

\subsection{Evaluation of Concepts}

\subsubsection{Quantitative Evaluation via Importance Score.}
\begin{figure}[H]
    \centering    \includegraphics[width=\linewidth]{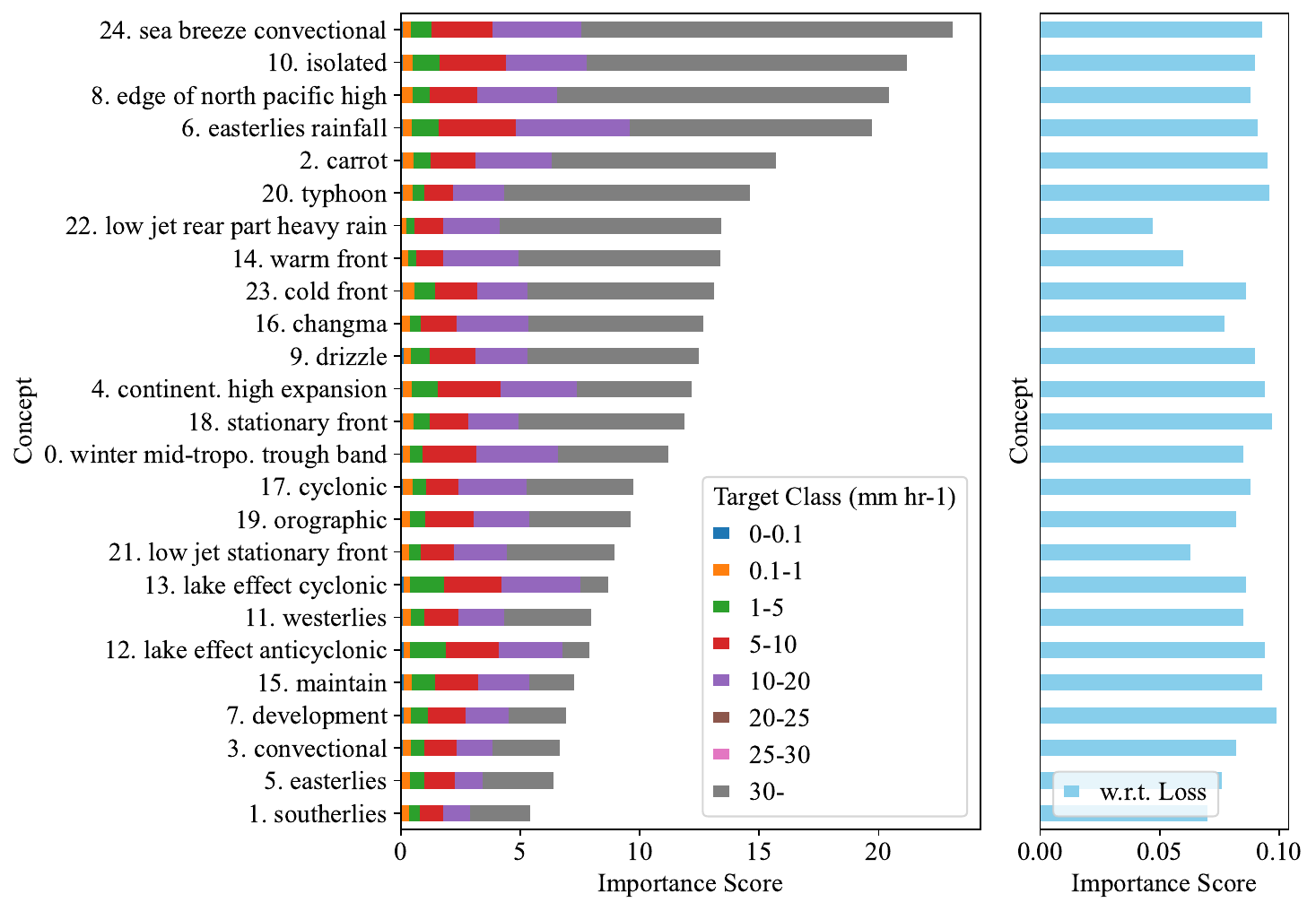}
    \caption{The importance score of concept activation vectors for each concept. The left panel displays the scores with respect to each target class, while the right panel represents the scores with respect to the loss value, i.e., encompassing all classes. The numbers preceding the concept labels indicate the label indices.} \label{fig:score}
\end{figure}

% However, their samples are annotated at different stages of generation so that the identified} CAVs show low importance scores for the overall target classes, even though their scores with respect to loss are relatively high.
% Note that only 25 out of 63 concepts are shown to conserve space. All scores are provided in Appendix~\ref{appx:totalscores}.}
We evaluate the quality of concept activation vectors (CAVs) by applying the importance score (in Section~\ref{method}.\ref{method:metric}) with the concept label dataset (in Section~\ref{method}.\ref{method:label}). Since no discussion on the good scores has yet been reported, we only make relative comparisons.
As shown in Fig.~\ref{fig:score}, the identified concepts are more sensitive to the over 30 mm hr-1 heavy rainfall class. In particular, concepts typically associated with heavy rainfall have high importance, such as \texttt{sea breeze convectional}, \texttt{isolated thunderstorm}, \texttt{edge of north pacific high}, \texttt{easterlies rainfall}, \texttt{carrot (tapering cloud)}
    \footnote{A carrot-shaped cloud or a tapering cloud is a convective cloud system with a narrow, triangular shape at its southwest end, often characterized by carrot-shaped cloud structure thinning toward the windward direction~\citep{jma2002analysis,toyoda1999midtropospheric}. It consists of cumulonimbus clouds extending from windward to leeward sides and is typically associated with heavy rain. Tapering cloud often have a lifespan of less than 10 hours~\citep{jma2002analysis}.}
, \texttt{typhoon}, \texttt{low level jet stream rear part of heavy rainfall}, fronts and \texttt{changma}
    \footnote{Changma refers to a meteorological phenomenon caused by the stationary front formed within the East Asian monsoon system~\citep{lee2017long, seo2011new}.}
. On the other hand, the importance is low for movement-related concepts such as \texttt{southerlies}, \texttt{easterlies}, and \texttt{maintain}.
Although \texttt{Convectional} and \texttt{development} are semantically important for rainfall generation, their CAVs are not sensitive to each target class during model prediction. 
This is because their samples are annotated at different stages of generation and the averaged activation vector could be not sensitive to model prediction especially with respect to individual target class.
It should be noted that 25 out of 63 concepts are shown in the main text to conserve space. The remaining scores are provided in Appendix~\ref{appx:totalscores}.

As shown in Fig.~\ref{fig:score}, the order of the scores of individual concept labels is inconsistent across the left and right panels 
This is because the loss function is based on modified F1 score~\ref{eq:modified_f1}
   \footnote{The objective function of the target model is specifically designed to prioritize the detection of heavy rainfall intensity by incorporating the accumulated target class.},
covering the entire set of output classes and suppressing the effect of over- and underestimated predictions.
As results, we can identify the model is overfitted to the higher rainfall intensity than lighter rainfall due to the behavior of the objective function, and target classes of 22-25 and 25-30 mm hr -1 is neglected throughout the concepts.
This specific scoring can serve as informative debugging guidance for modeling engineers. Forecasters can be provided the importance scores as a measure of confidence in the concept explanation.

Although the importance score with respect to the loss effectively illustrates the behavior of the model's loss surface, the objective function can vary across modeling designs. For conventional use, comparing class-wise importance scores can make the results more interpretable for humans. In our case, class-wise scores allow the users to explore the effect of concepts for the predictions in the target classes of interest.

Hence, compared to the score with respect to the loss, the method using a wrapper function to aggregate logits separately by each target class has the advantage for the model of forecasting rainfall intensity across different intervals, allowing exploration of which concepts are important for prediction in the target classes of interest.

\subsubsection{Qualitative Evaluation via Perturbation Test.}
\begin{figure}[H]
    \centering
    \includegraphics[width=\linewidth]{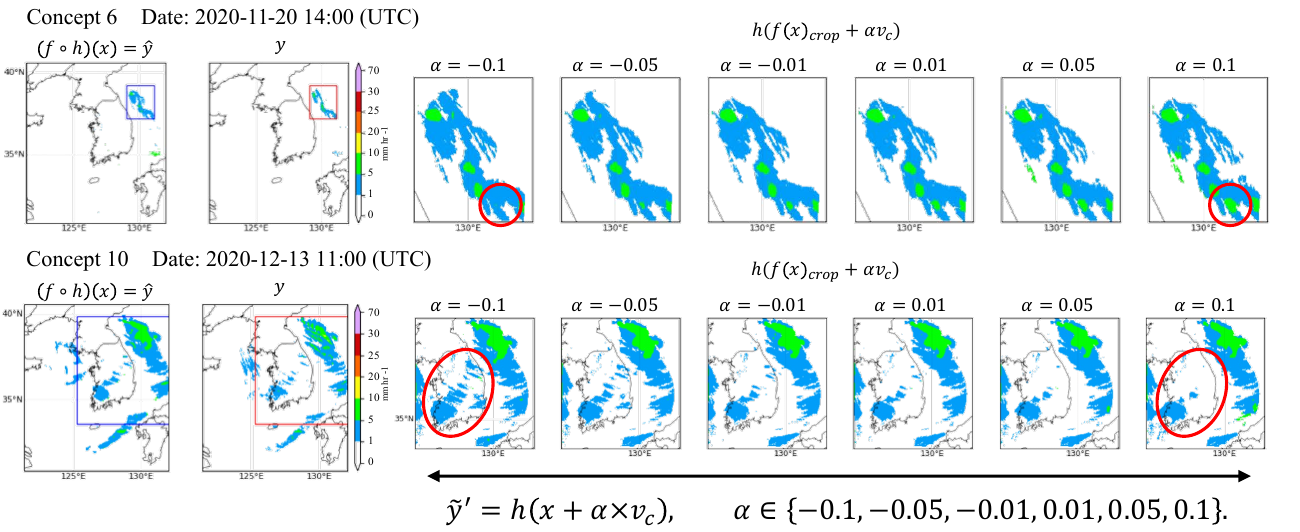}
    \caption{Perturbation test of concept activation vectors. $\hat{y}, y,$ and $\tilde{y}'$ denote one hour ahead prediction, the ground truth, and the perturbed predictions, respectively. ``Examples of concept 6 (\texttt{easterlies rainfall}) and concept 10 (\texttt{isolated}) illustrate the nonlinear development effect on future predictions while retaining their underlying mechanisms.''}
    \label{fig:perturbation}
\end{figure}

Fig.~\ref{fig:perturbation} demonstrates the effect of performing perturbations in the direction of CAVs. We can visually identify nonlinear development and dissipation patterns, indicating that the target model captures nonlinear rainfall mechanisms in its feature representation space.
The increase or decrease in the CAV values for the \texttt{easterlies rainfall} concept represents the expansion or contraction of a 5 mm/hr heavier rainfall area in the southern part of the precipitation system when predicting an example of easterlies rainfall (dated November 20, 2020, at 14:00 UTC).
In another case, for \texttt{isolated} concept, the increase or decrease in CAV values shows the development or dissipation of a 1-5 mm hr -1 scattered light rainfall area when predicting an example of scattered rainfall (dated December 13, 2020, at 11:00 UTC). This module can assist forecasters who are investigating when the effect of a specific rainfall mechanism is amplified or diminished.

\subsection{Concept Prober as a Tool for Model Debugging Guidance}

\begin{figure}[H]
    \centering
    \includegraphics[width=0.85\linewidth]{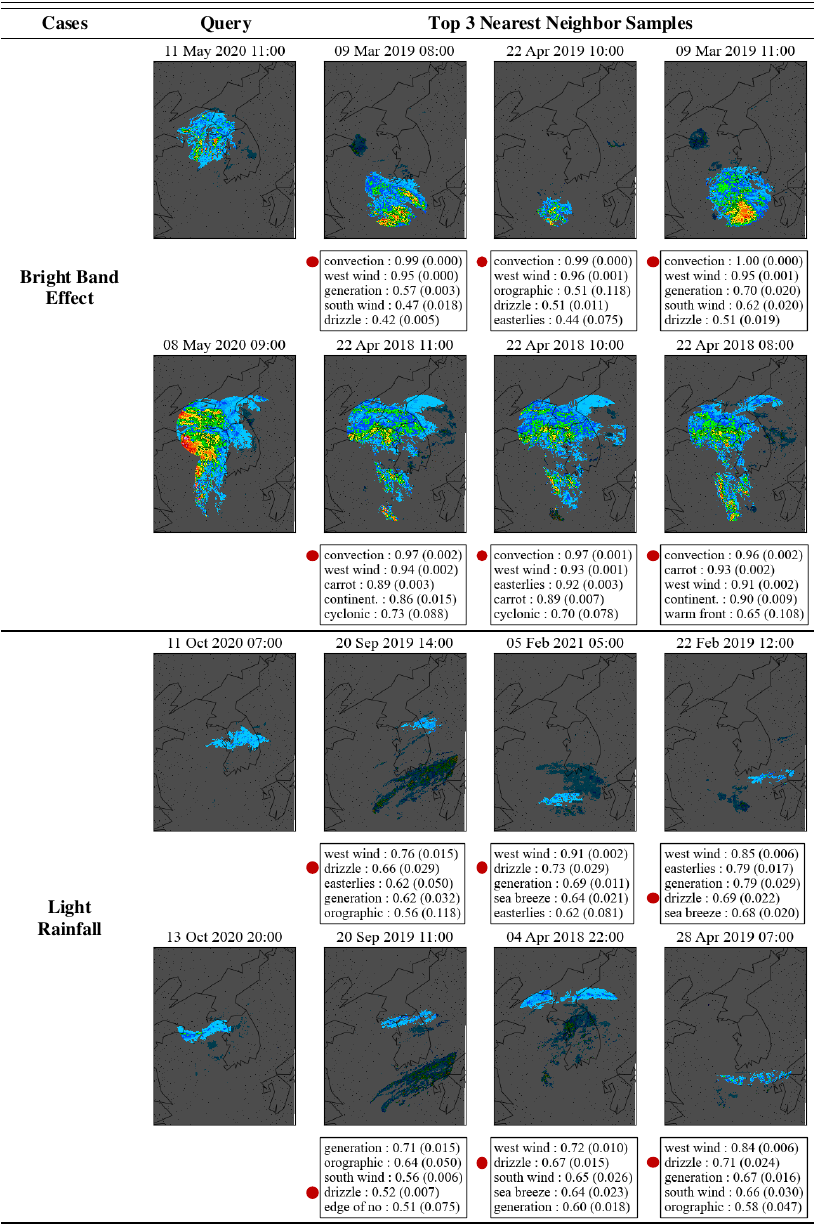}
    \caption{Probabilities and uncertainties of concept probers on bright band effect and light rainfall cases. Each concept explanation is accompanied by its predictive probability (shown as the left number) and its uncertainty (in parentheses).}
    \label{fig:uncertainty}
\end{figure}

We investigate cases of exception detection to provide insights for model debugging. In this context, uncertainty can serve as a valuable explanation tool for users. Specifically, we compute epistemic uncertainty using ensemble-based linear probers within a 5-fold cross-validation splitting strategy. This involves calculating the variance of the predictive probabilities generated by five trained linear classifiers, as illustrated in Fig.~\ref{fig:uncertainty}.

Model debugging for engineers can be approached in two primary steps: data collection and model development. Accordingly, we hypothesize the following:
1) measurement errors in radar data can be identified using concept probers, and 
2) insufficient model representations can also be diagnosed using concept probers.
To evaluate these hypotheses, we analyze two specific cases:
1) bright band samples caused by measurement errors, and 
2) light rainfall cases to investigate whether the model is predisposed to overestimation.
These examples are derived from annotated documents provided by forecaster reports, as introduced in Section~\ref{method}.\ref{method:label}.

In Fig.~\ref{fig:uncertainty}, the first two rows correspond to bright band examples, and the next two rows represent light rainfall cases such as \texttt{drizzle}.
For the first case, concept probers tend to classify bright band examples as rainfall driven by the \texttt{convection} mechanism with near 100\% certainty. While it is not possible to explicitly train the prober for the bright band effect due to data limitations, these cases suggest potential overestimation caused by the bright band effect when concept probers consistently identify the concept as \texttt{convection} with almost zero uncertainty.
In contrast, the prober's performance in distinguishing light rainfall cases is relatively low and exhibits high uncertainty. We posit that this is likely due to the model being biased towards overestimation and having insufficient internal representations for light rainfall concepts such as \texttt{drizzle}, \textit{isolated}, or \texttt{dissipation}, leading to confusion in detecting light rainfall concept. 
The implication above indicates that uncertainty information derived from concept probers can offer significant support for effective model debugging.

\subsection{User Interface}
\begin{figure}[H]
\centering
\includegraphics[width=0.6\linewidth]{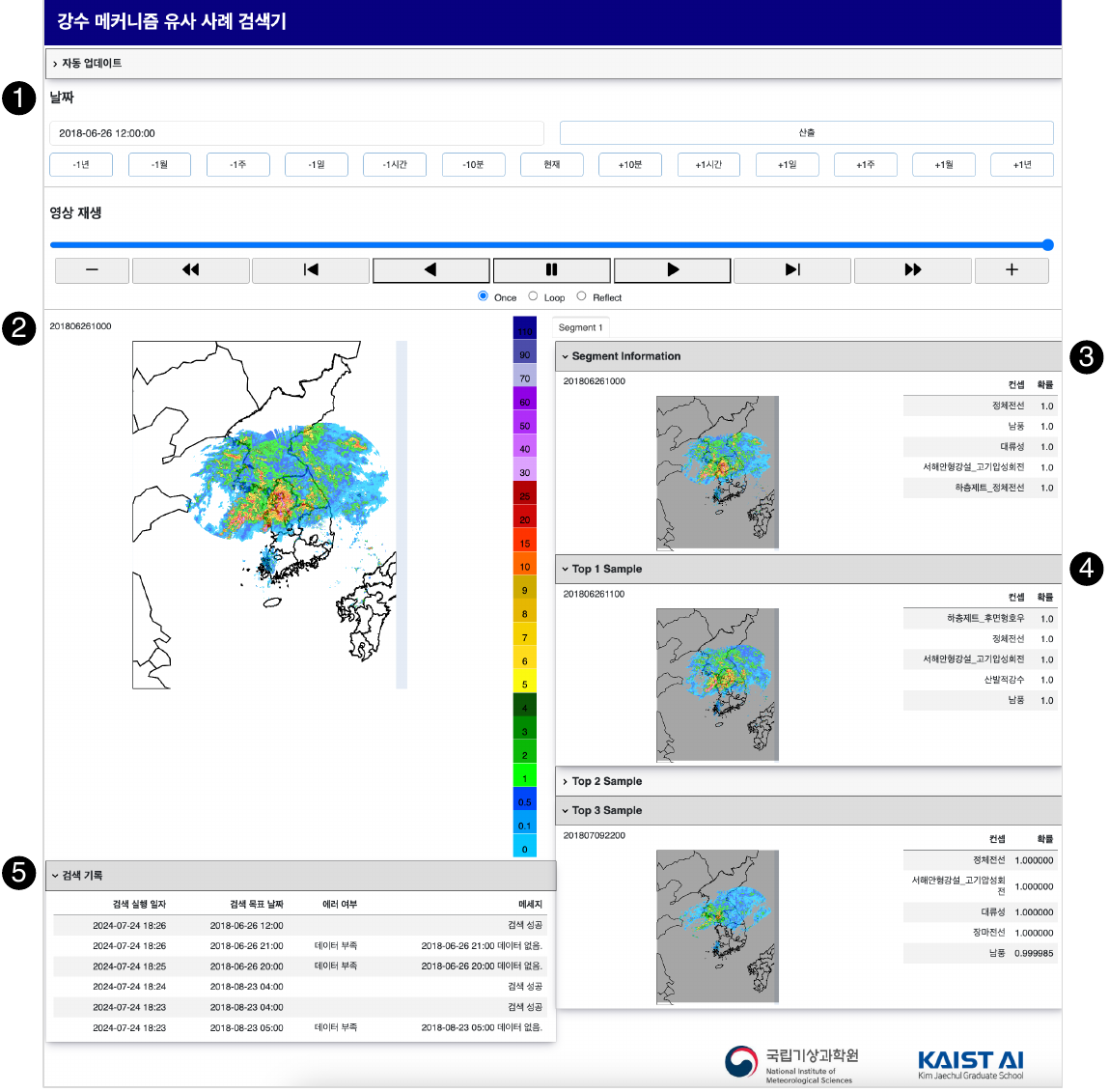}
\caption{User interface of example-based probabilistic concept explanation (in Korean). The numerics in black circles denote the functions of user interface. The detailed description is provided in Table~\ref{tab:ui}.}\label{fig:ui}
\end{figure}

\begin{figure}[H]
\centering
\includegraphics[width=\linewidth]{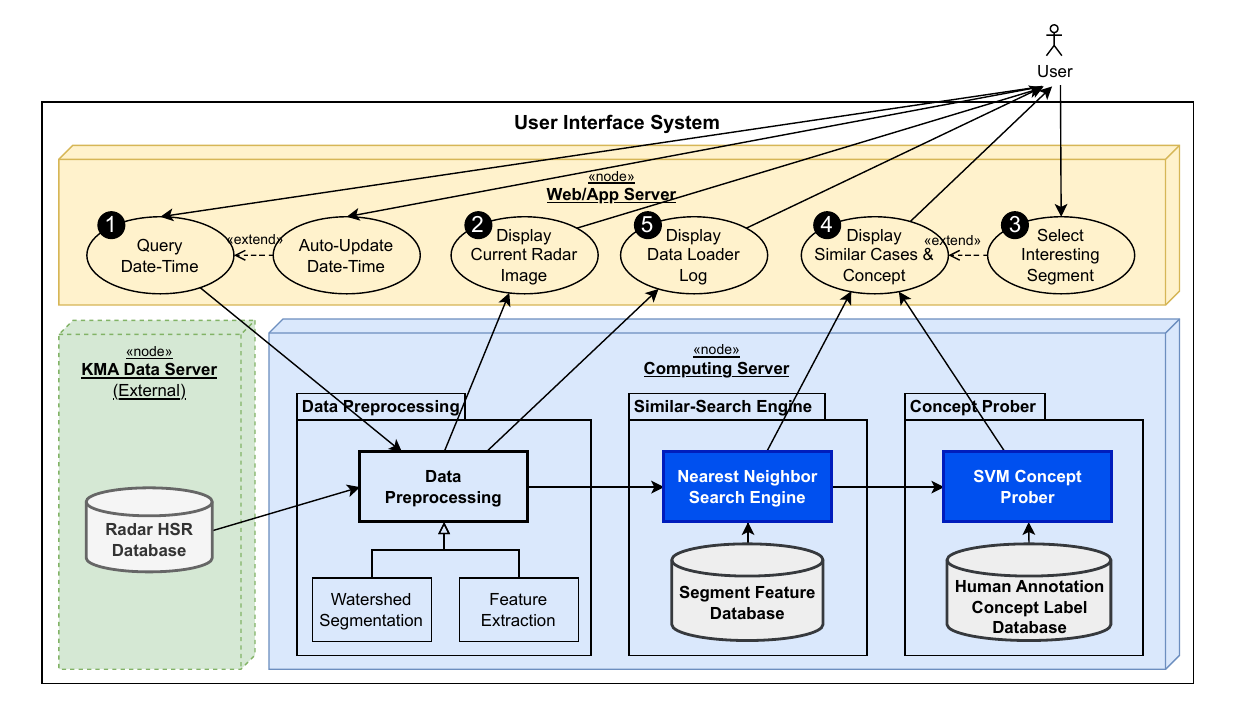}
\caption{A use case diagram illustrating the operations of our framework and associated UI. Note that the radar database is outside the system. The numerics in black circles denote the functions of user interface. The detailed description is provided in Table~\ref{tab:ui}.}\label{fig:usecase}
\end{figure}

Incorporating the explanations from the neighbor search engine, the proposed user interface (UI) consists of five components: 1) query date selector, 2) main radar data display, 3) search logs for debugging, 4) precipitation segment display, and 5) neighbor search engine result display (Fig.~\ref{fig:ui}).
The UI functionalities have been designed to balance the number of steps required to generate output and user's control over the generation process (Table~\ref{tab:ui}).
The UI service is currently at a ready-to-deploy state in the intranet of Synoptic Chart Analysis Comprehensive Portal provided by KMA. 
A use case is designed as shown in Fig.~\ref{fig:usecase}.
We build the UI via \texttt{Panel}
    \footnote{\url{https://panel.holoviz.org/reference/panes/Plotly.html}}
, an open-source library for web application development, and \texttt{Plotly}
    \footnote{\url{https://plotly.com/python/}}
, an open-source library for user interactive visualization.
% http://cht.kma.go.kr/

\vspace{-0.4in}

\begin{table}[H]
    \centering
    \caption{The functions and components of user interface.}
    \label{tab:ui}
    % \scriptsize
    % \small
    \fontsize{12}{10}\selectfont 
    \begin{tabularx}{\linewidth}{>{\hsize=0.21\hsize}X
                                 >{\hsize=0.29\hsize}X
                                 >{\hsize=0.55\hsize}X}
    \hline\hline
    \textbf{Functions} & \textbf{Components} & \textbf{Descriptions} \\\hline
    \multirow{6}{=}{\makecell[l]{\circledtext*{1} Target date \\selection \& \\output settings}}
        & Date selection widget 
        & Select query date for search \\\cline{2-3}
        & Date change button
        & Change target query date by 10 minutes, 1 day, 1 week, 1 month, or 1 year intervals \\\cline{2-3}
        & Auto update widget
        & Choose between automatic and manual updates\\\cline{2-3}
        & Output widget    
        & Search query dates and output similar samples\\\cline{2-3}
        & Animation player widget
        & Play animation of sequential time points of radar\\\hline
    \circledtext*{2} Display input
        & Input radar data panel
        & Display query radar data on the map\\\hline
    \multirow{2}{=}{\circledtext*{3} Select input segments}
        & Radar data panel
        & Display similar samples from the query data\\\cline{2-3}
        & Concept table
        & Display the top 5 concepts of the similar sample\\\hline
    \multirow{2}{=}{\circledtext*{4} Display similar cases}
        & Radar data panel
        & Display top 3 similar cases\\\cline{2-3}
        & Concept table
        & Display top 5 concepts of similar cases\\\hline
    \circledtext*{5} Search log
        & Search log table
        & Record past search results(search time, target query date, error/success message, etc.)\\\hline
    \end{tabularx}
\end{table}

\section{Conclusion}\label{conclusion}

We develop an example-based concept explanation framework to provide an easily approachable XAI tool for practitioners in meteorological operations. We create a rainfall mechanism concept dataset from domain materials and adopt supervised concept extraction methods to identify rainfall mechanisms from the internal representational space of trained DNN model. To search for conceptually similar cases from high-resolution images, we design a nearest neighbor search engine that incorporates principal neurons selection-based dimensionality reduction in the feature space for the computational efficiency. The importance of concepts with respect to the target class is computed by adapting the existing input attribution evaluation metrics for regular classifiers to our segmentation models through output wrapper functions. This procedure of our search engine identifies the nearest neighbors in the model's internal feature space as examples that share conceptually similar rainfall mechanisms with the query sample. We find that the concept probers can distinguish nonlinear development and dissipation mechanisms captured by the target model (related to the first research question) and can identify semantically meaningful meteorological attributes, aiding users evaluate whether patterns that are aligned with domain knowledge are reflected in the target model's inference process. (related to the second question). In light of application, If the internal model's representational space is well-trained, this framework also functions as a rainfall type classifier for unseen query data. We also demonstrate that this framework can function as a model debugging tool. As a collaborative effort with domestic forecasters and subject matter experts, we design a user interface framework to facilitate communication with domain users during the development of the XAI framework. This framework provides user-friendly explanations for AI models in meteorology, enhancing their trustworthiness with the ultimate goal of making them a viable alternative to NWP models in practice.

There are several directions that could be considered in the future. First, our label dataset could be improved through expansion, quality control, or augmentation with concepts extracted in an unsupervised manner. On the one hand, our labeled dataset is currently limited to precipitation phenomena observed in the Korean Peninsula. We may enhance the quality of the dataset by including new labels such as hail, snowfall, or rainfall elevation. We may extend to other regions or scales, or use a different model to construct a richer feature representation space that captures more informative concepts. Another aspect to consider is improving the overall quality of the data labels. Due to the characteristics of the source material, the quality of labeled data depends on the quality of the annotators, such as individual forecasters or material authors. We may address this concern by adopting a labeling system with voting mechanism using the number of votes as confidence in the chosen label. On the other hand, given that the current feature space seems to capture conceptual patterns aligned with domain knowledge, we may be able to extract concepts directly from the feature vectors to add as labels. Second, the framework may be extended to different categories of models such as generative models. Given that feature space analysis is often performed for generative models, it seems like an appropriate choice as the next step in research. Finally, the type of explanations may be extended to causality with other variables. Our current algorithm is designed to match our target model in input and output, which limits our explanations to be completely radar data-based which is the consequence of precipitation process. This limitation constrains its ability to address the causal mechanisms with other variables underlying precipitation systems, an aspect that domain users may be interested in. By incorporating variables including thermal instability or convergence at different altitudes, this framework could facilitate the extraction of causal information within DNNs.

\clearpage
%%%%%%%%%%%%%%%%%%%%%%%%%%%%%%%%%%%%%%%%%%%%%%%%%%%%%%%%%%%%%%%%%%%%%
% ACKNOWLEDGMENTS
%%%%%%%%%%%%%%%%%%%%%%%%%%%%%%%%%%%%%%%%%%%%%%%%%%%%%%%%%%%%%%%%%%%%%
\acknowledgments
This work was supported by the Korea Meteorological Administration and Korean National Institute of Meteorological Sciences under grant agreement No. KMA2021-00123 (Developing Intelligent Assistant Technology and Its Application for Weather Forecasting Process), and from the Korean Institute of Information \& Communications Technology Planning \& Evaluation and the Korean Ministry of Science and ICT under grant agreement No. RS-2019-II190075 (Artificial Intelligence Graduate School Program(KAIST)) and No. RS-2022-II220984 (Development of Artificial Intelligence Technology for Personalized Plug-and-Play Explanation and Verification of Explanation).
%  Keep acknowledgments (note correct spelling: no ``e'' between the ``g'' and
% ``m'') as brief as possible. In general, acknowledge only direct help in
%  writing or research. Financial support (e.g., grant numbers) for the work done, 
%  for an author, or for the laboratory where the work was performed must be 
%  acknowledged here rather than as footnotes to the title or to an author's name.
%  Contribution numbers (if the work has been published by the author's institution 
%  or organization) should be placed in the acknowledgments rather than as 
%  footnotes to the title or to an author's name.

%%%%%%%%%%%%%%%%%%%%%%%%%%%%%%%%%%%%%%%%%%%%%%%%%%%%%%%%%%%%%%%%%%%%%
% DATA AVAILABILITY STATEMENT
%%%%%%%%%%%%%%%%%%%%%%%%%%%%%%%%%%%%%%%%%%%%%%%%%%%%%%%%%%%%%%%%%%%%%
% 
%
\datastatement
We use radar hybrid surface rainfall (HSR) observation which is developed within KMA's Weather Radar Center and is publicly available in the Korean National Climate Data Center (https://data.kma.go.kr/data/rmt/rmtList.do?code=11pgmNo=62 in Korean, https://data.kma.go.kr/resources/html/en/aowdp.html in English.)
Our code is available in the Figshare repository (doi:10.6084/m9.figshare.27993743).
%  The data availability statement is where authors should describe how the data underlying 
%  the findings within the article can be accessed and reused. Authors should attempt to 
%  provide unrestricted access to all data and materials underlying reported findings. 
%  If data access is restricted, authors must mention this in the statement. See
%  {http://www.ametsoc.org/PubsDataPolicy} for more info.

%%%%%%%%%%%%%%%%%%%%%%%%%%%%%%%%%%%%%%%%%%%%%%%%%%%%%%%%%%%%%%%%%%%%%
% APPENDIXES
%%%%%%%%%%%%%%%%%%%%%%%%%%%%%%%%%%%%%%%%%%%%%%%%%%%%%%%%%%%%%%%%%%%%%
%
%% If only one appendix, use
\clearpage
% \appendix

%% If more than one appendix, use \appendix[<letter>], e.g.,

\appendix[A]
\appendixtitle{Data and Model}

\section{Model Overview}\label{appx:data}
The target model is an unpublished variant of DeepRaNE~\citep{ko2022effective}, provided by the National Institute of Meteorological Sciences(NIMS). It features a convolution-based denoising autoencoder combined with a U-Net structure for pixel-wise rainfall intensity classification. This model classifies precipitation into eight categories: 0-0.1, 0.1-1, 1-5, 5-10, 10-20, 20-25, 25-30, and 30 mm hr -1. Predictions are made at one-hour intervals with a lead time ranging from 1 to 6 hours.
The model architecture is described in Table~\ref{tab:model}.
\renewcommand{\arraystretch}{1}
\begin{table}[H]
\centering
\fontsize{10}{7}\selectfont 
\caption{The precipitation forecast model architecture. It consists of denoising autoencoder (DAE) and U-Net.}
\label{tab:model}
\begin{tabularx}{\textwidth}{p{0.12\textwidth}p{0.14\textwidth}p{0.14\textwidth}p{0.55\textwidth}}
\hline\hline
\textbf{Layer Name} & \textbf{Input Shape} & \textbf{Output Shape} & \textbf{Operation Details} \\ \hline

\multicolumn{4}{l}{\textbf{1. DAE}} \\ \hline
Encoder
& $[13, 1024, 1152]$ 
& $[16, 513, 577]$ 
& Conv2d($3\times3$, pad=2), ReLU, BatchNorm2d, MaxPool2d($2\times2$) \\ 
& $[16, 513, 577]$ 
& $[32, 513, 577]$ 
& Conv2d($3\times3$, pad=1), ReLU, BatchNorm2d \\
& $[32, 513, 577]$ 
& $[64, 257, 290]$ 
& Conv2d($3\times3$, pad=2), ReLU, BatchNorm2d, MaxPool2d($2\times2$) \\
& $[64, 257, 290]$ 
& $[128, 257, 290]$ 
& Conv2d($3\times3$, pad=1), ReLU \\ \hline
Decoder
& $[128, 257, 290]$ 
& $[64, 514, 580]$ 
& ConvTranspose2d($3\times3$, stride=2, pad=1, output\_pad=1), ReLU, BatchNorm2d\\ 
& $[64, 514, 580]$ 
& $[16, 514, 580]$ 
& ConvTranspose2d($3\times3$, pad=1), ReLU, BatchNorm2d \\
& $[16, 514, 580]$ 
& $[8, 1028, 1160]$ 
& ConvTranspose2d($3\times3$, stride=2, pad=1, output\_pad=1), ReLU \\ \hline

\multicolumn{4}{l}{\textbf{2. U-Net}} \\ \hline
 InitialConv
 & $[13, 1024, 1152]$ 
 & $[32, 1024, 1152]$ 
 & Conv2d($3\times3$ kernel, pad=1 ) \\ \hline
 SecondConv
 & $[32, 1024, 1152]$ 
 & $[32, 1024, 1152]$ 
 & Conv2d($3\times3$ kernel, pad=1)  \\ \hline
 DownSample
 & $[32, 1024, 1152]$ 
 & $[64, 512, 576]$ 
 & MaxPool2d($2\times2$), 2 Conv2d($3\times3$, pad=1) \\ 
 & $[64, 512, 576]$ 
 & $[128, 256, 288]$ 
 & MaxPool2d($2\times2$), 2 Conv2d($3\times3$, pad=1) \\ 
 & $[128, 256, 288]$ 
 & $[256, 128, 144]$ 
 & MaxPool2d($2\times2$), 2 Conv2d($3\times3$, pad=1) \\ 
 & $[256, 128, 144]$ 
 & $[512, 64, 72]$ 
 & MaxPool2d($2\times2$), 2 Conv2d($3\times3$, pad=1) \\ 
 & $[512, 64, 72]$ 
 & $[1024, 32, 36]$ 
 & MaxPool2d($2\times2$), 2 Conv2d($3\times3$, pad=1) \\ \hline
 
 UpSample
 & $[1024, 32, 36]$ 
 & $[512, 64, 72]$ 
 & ConvTranspose2d($2\times2$, stride=2, pad=0), Concat, 2 Conv2d($3\times3$, pad=1) \\ 
 & $[512, 64, 72]$ 
 & $[256, 128, 144]$ 
 & ConvTranspose2d($2\times2$, stride=2, pad=0), Concat, 2 Conv2d($3\times3$, pad=1) \\ 
 & $[256, 128, 144]$ 
 & $[128, 256, 288]$ 
 & ConvTranspose2d($2\times2$, stride=2, pad=0), Concat, 2 Conv2d($3\times3$, pad=1) \\ 
 & $[128, 256, 288]$ 
 & $[64, 512, 576]$ 
 & ConvTranspose2d($2\times2$, stride=2, pad=0), Concat, 2 Conv2d($3\times3$, pad=1)  \\
 & $[64, 512, 576]$ 
 & $[32, 1024, 1152]$ 
 & ConvTranspose2d($2\times2$, stride=2, pad=0), Concat, 2 Conv2d($3\times3$, pad=1) \\ \hline
 LastConv
 & $[32, 1024, 1152]$ 
 & $[1, 1024, 1152]$ 
 & Conv2d($3\times3$, pad=1) \\ \hline
\end{tabularx}
\end{table}

The objective function is a specialized F1 score designed to focu on heavy rainfall:
\begin{equation}
\begin{aligned}
\text { Modified F1 }=\frac{1}{7} & \left( \frac{\operatorname{Hit}_{0.1 \mathrm{mm} / \mathrm{h} \text { over }}}{\text { Hit }_{0.1 \mathrm{mm} / \mathrm{h} \text { over }}+\frac{1}{2}\left(\mathrm{Miss}_{0.1 \mathrm{mm} / \mathrm{h} \text { over }}+\text { FalseAlarm }_{0.1 \mathrm{mm} / \mathrm{h} \text { over }}\right)} \right.\\
& + \frac{\operatorname{Hit}_{1 \mathrm{mm} / \mathrm{h} \text { over }}}{\text { Hit }_{1 \mathrm{mm} / \mathrm{h} \text { over }}+\frac{1}{2}\left(\mathrm{Miss}_{1 \mathrm{mm} / \mathrm{h} \text { over }}+\text { FalseAlarm }_{1 \mathrm{mm} / \mathrm{h} \text { over }}\right)} + \cdots \\
& \left.+ \frac{\mathrm{Hit}_{10 \mathrm{mm} / \mathrm{h} \text { over }}}{\mathrm{Hit}_{10 \mathrm{mm} / \mathrm{h} \text { over }}+\frac{1}{2}\left(\text { Miss }_{10 \mathrm{mm} / \mathrm{h} \text { over }}+ \text { FalseAlarm }_{10 \mathrm{mm} / \mathrm{h} \text { over }}\right)}\right)
\end{aligned}
\label{eq:modified_f1}
\end{equation}

\section{Literature for Rainfall Mechanism Classification}\label{appendix:reference}

\begin{figure}[H]
    \centering
    \includegraphics[width=\linewidth]{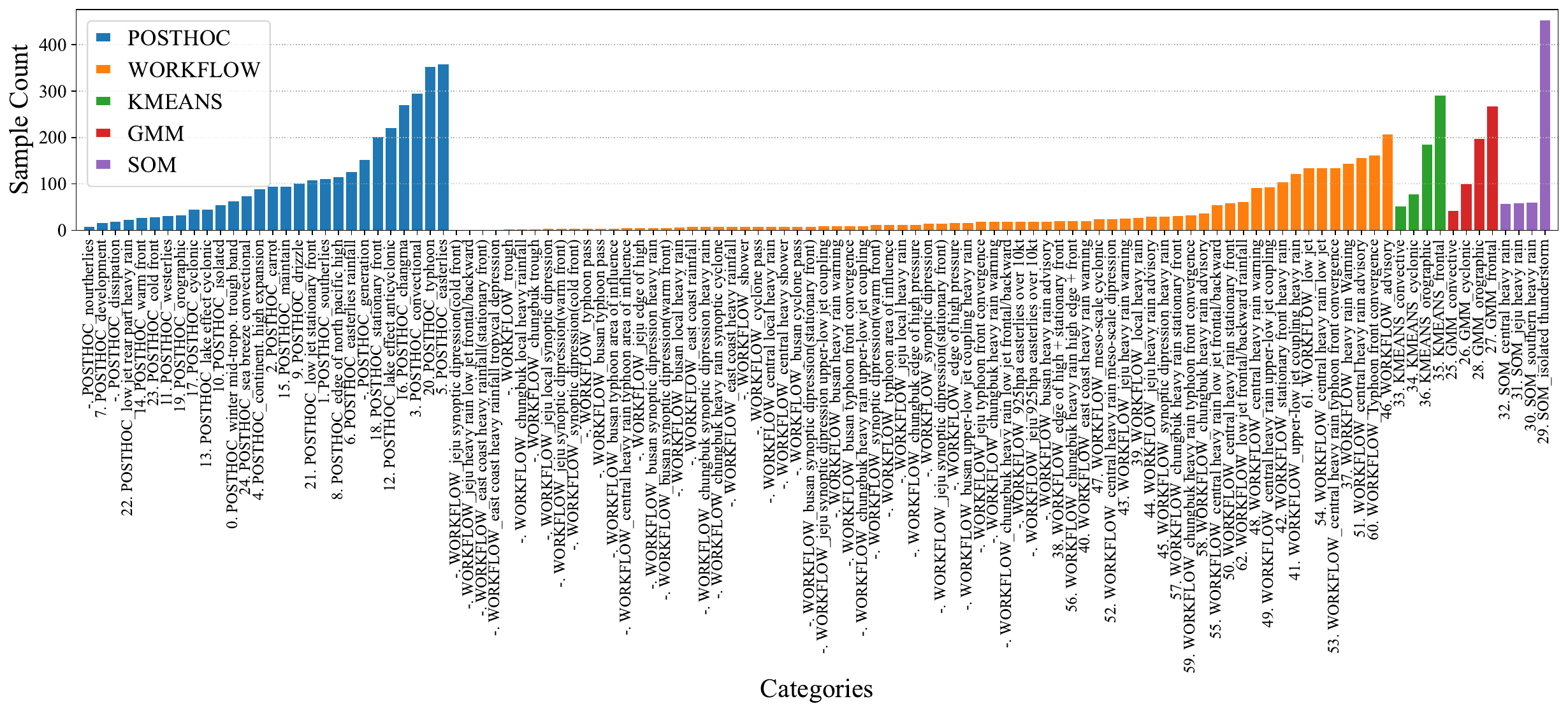}
    \caption{The number of samples of human-annotated concept labels. 
    `POSTHOC' denotes the labels from post-hoc forecast analysis reports and case studies from the references of the first four rows in Table~\ref{tab:reference}. 
    `WORKFLOW' indicates the labels based on confidential materials provided by NIMS.
    `KMEANS' and `GMM' represent the labels annotated based on the fifth row in Table~\ref{tab:reference}.
    `SOM' denotes the labels annotated from the last three rows, including \cite{jo2020classification} and \cite{park2021diverse} in the Table~\ref{tab:reference}.
    The numerics in front of individual X-axis labels denote the label index used in this paper. 
    The null index of `-' indicates the label whose number of samples is below 20.}
    \label{fig:numsample}
\end{figure}

The review of previous studies on precipitation mechanism classification conducted by collaborating institutions are provided in Table~\ref{tab:reference}. The first four rows are case studies for classifying precipitation types. We use cases that fall within the scope of our study's training data (2018 - 2021 inclusive.) as additional materials for concept labels based on the forecasters' reports of post hoc weather prediction. 
The next four rows report research using clustering models such as self-organizing map (SOM), K-Means, and Gaussian mixture model (GMM) primarily focused on the cases with heavy precipitation of 10 mm hr-1 or 30 mm hr-1 and above. We use these studies to build additional label sets. 
The total number of the samples is 7,343, comprising 3,147 for `POSTHOC', 2,357 for `WORKFLOW', 604 for `KMEANS', 606 for `GMM', and 629 for `SOM'.
The number of samples of human-annotated concept labels are presented in Fig.~\ref{fig:numsample}.

% \begin{landscape}
\begin{table}[H]
\caption{References related to the rainfall mechanism classification.}\label{tab:reference}
\centering
% \begin{tabular}{llllll}
\scalebox{0.95}{
    % \scriptsize
    \fontsize{10}{8}\selectfont 
    \begin{tabularx}{1\linewidth}{>{\hsize=0.2\hsize}X
                                 >{\hsize=0.11\hsize}X
                                 >{\hsize=0.09\hsize}X
                                 >{\hsize=0.11\hsize}X
                                 >{\hsize=0.13\hsize}X
                                 >{\hsize=0.36\hsize}X}
    \hline
    \textbf{Title} & \textbf{Author} & \textbf{Date} & \textbf{Method} & \textbf{Data} & \textbf{Category(\# of cases)}
    \\\hline
    Development of Weather-AI Data Preprocessing Technology(in Korean) & Natl. Inst. of Met. Sciences (NIMS) 
    & 2022 
    & Case study
    & 2021-2022, Jeju region, weather chart 
    & Low pressure passage(12), indirect effect of low pressure(3), Changma front(4), mesoscale convective(3), air mass changing snowfall(12)
    \\\hline
    Guidance on Satellite-Based Objective Cloud Analysis Technology(in Korean) 
    & NIMS 
    & 2022 
    & Case study
    & 2013-2017, weather chart
    & Low pressure passage(2), frontal low pressure(2), lower-level jet(3), Changma front(2), mesoscale convective(2), air mass changing snowfall(7)
    \\\hline
    Forecaster's Handbook Series 2: Comprehensive Concept Model of Heavy Rain(in Korean) 
    & Forecast. Tech. Team, Korea Met. Admin. (KMA) 
    & 2010 
    & Case study
    & 2002-2010, weather chart
    & Thickness in front of lower-level jet(3), thickness behind lower-level jet(5), convergence in front of typhoon(5), tropical depression(4), direct effect zone of typhoon(5), East Coast heavy rainfall(1)
    \\\hline
    Practical Forecasting Techniques - Utilization and Definition of Essential Forecast Elements(in Korean) 
    & Forecast. Technology Team, KMA 
    & 2014 
    & Case study
    & 2001-2011, weather chart
    & Upper and lower-level jet coupling(Changma and second rainy season)(27), convergence in front of typhoon(3), typhoon(5), isolated heavy rainfall(8), developed low pressure(1)
    \\\hline
    Development of Weather-AI Data Preprocessing Technology I(in Korean) 
    & Environ. Pred. Res., Sejong Univ., NIMS 
    & 2022 
    & Model-Based: SOM, K-Means, GMM
    & 2005-2022, 1h cumul. precip. ERA5 Reanal. II 
    & Low pressure, convective, orographical, fronts, others
    \\\hline
    Development of Weather-AI Data Preprocessing Technology II(in Korean) & Seoul National Univ., NIMS 
    & 2022 
    & Model-Based: SOM
    & 2005-2017, Jun, Jul, Aug, and Sep (JJAS)
    & Central region, isolated, southern region, jeju region
    \\\hline
    Classification of localized heavy rainfall events in South Korea 
    & Jo, et al.,\citep{jo2020classification} 
    & 2019 
    & Model-Based: SOM, K-Means 
    & 2005-2017, JJAS
    & Front-related band in central region, southern region, isolated heavy rainfall
    \\\hline
    Diverse Synoptic Weather Patterns of Warm-Season Heavy Rainfall Events in South Korea 
    & Park, et al.\citep{park2021diverse} 
    & 2021 
    & Model-Based: SOM
    & 2005-2017, JJAS, ASOS, ERA-Interim 1.5$^{\circ}$
    & Quasi-stationary frontal boundary between low and high, extratropical cyclone in Eastern China, local disturbances at the edge of the North Pacific High, moisture pathway between continental high and oceanic high.
    \\\hline
    \end{tabularx}
}
\end{table}

\appendix[D] 
\appendixtitle{Experimental Settings}

\section{Appropriate Number of Dimensions for Relaxed Decision Region}\label{appx:dimensionality}

We set the hyperparameter of the principal neuron component-based neighbor search engine to 300 in Section~\ref{method}.\ref{method:framework}.\ref{method:nn}. We select this value by comparing hyperparameter settings of {15, 100, 300, 1000}. The default setting in the literature~\citep{chang2024understanding} is 15. 

\textit{Runtime} is evaluated for each individual query sample, with relative performance assessed using \textit{Precision@k} which represents the proportion of correct results within the top k nearest neighbors, determined using human-annotated labels (refer to Section~\ref{method}.\ref{method:label}): $ Precision @ k = \frac{|\text{correct samples among } k \text{ results}|}{k} $.
The experiment is conducted on an Intel Xeon Gold 6342 CPU @ 2.8GHz with 96 logical cores. The random seed is fixed at 42.

As shown in Table~\ref{tab:dimensionality}, the results indicate a trade-off between performance and runtime across different number of dimensions. We adopt a dimensionality of 300 based on this trade-off.

\begin{table}[H]
\setlength{\tabcolsep}{5pt}
\caption{Comparison of runtime and precision across hyperparameters for the number of dimensions in the principal neuron component-based neighbor search engine (PC-NSE).}
\centering

\begin{tabular}{llcccc}
\hline\hline
Embedding & \# Dim & Runtime (sec) & Precision@3 & Precision@5 & Precision@10 \\
\hline
$Z_{PC-NSE}$
    & 15
    & \textbf{1.40}
    & 0.3333 $\pm$ 0.1323 & 0.1771 $\pm$ 0.0866 & 0.1086 $\pm$ 0.0657 \\ 
$Z_{PC-NSE}$
    & 100 
    & 1.50
    & 0.3571 $\pm$ 0.1470 & 0.2686 $\pm$ 0.1228 &  0.1757 $\pm$ 0.0828 \\
$Z_{PC-NSE}$
    & 300
    & 1.50
    & 0.4667 $\pm$ 0.1795 & 0.3171 $\pm$ 0.1323 & 0.1943 $\pm$ 0.0916 \\
$Z_{PC-NSE}$
    & 1,000
    & 2.70
    & \textbf{0.5333 $\pm$ 0.1788} & \textbf{0.3686 $\pm$ 0.1412} & \textbf{0.2257 $\pm$ 0.1296} \\
    \hline
\end{tabular}

\label{tab:dimensionality}
\end{table}

\section{Appropriate Number of Samples for Each Concept}\label{appx:numsample}

\cite{kim2018interpretability} suggest that 10 to 20 images are enough to compute concept activation vectors (CAV) over all 1000 classes of ImageNet dataset, while~\cite{ghorbani2019towards} suggest 50 images for 100 subclasses. However, we find a trade-off between the number of target classes are the minimum number of samples: in our experimental case, 43 concept classes are used for a threshold of 50 samples, 63 concepts for 20 samples, and 82 concepts for 10 samples. 

To analyze the number of samples to compute CAV, we measure the importance score on the cases of 10, 20, and 50 samples as shown in Table~\ref{tab:numsample2}. Loss score is used to compute importance scores. Randomly chosen 50 samples are computed with random seed of 42 throughout the three candidates. We empirically do not find a trend in importance score quality. Therefore, we set the minimum number to 20 to cover a greater number of concepts. 

\begin{table}[H]
    \centering
    \caption{The importance scores on the number of samples for each concept class.}
    \begin{tabular}{cc}
         \hline\hline
         \# of Samples & Importance Score \\
         \hline
         10 & 0.098 $\pm$ 0.024 \\
         20 & 0.084 $\pm$ 0.013 \\
         50 & \textbf{0.115} $\pm$ \textbf{0.022}\\
         \hline
    \end{tabular}
    \label{tab:numsample2}
\end{table}

\section{Study on Weighting by Temporal Distance.}
One potential issue with nearest neighbor search is the selection of temporally close samples as a measure of conceptual similarity. As a possible solution, we design a weighting function for the temporal distance to query sample. In our dataset, the time intervals are uniformly set to 1-hour units, allowing a simple application of weights based on the magnitude of the time difference:

$$
w(t) = \frac{1}{(\epsilon+|\Delta t|)^2}
$$
where $\Delta t$ denotes the time difference between query point and another point, and $\epsilon$ represents a very small value (e.g., $1e-8$). We use Euclidean distance as distance function $d(x) = ||\mathcal{W}\circ\phi_l (x_{query}) - \mathcal{W}\circ\phi_l(x)||_2$, where $W$ is a Watershed segmentation and resizing function, and $\phi_l$ denotes the forward function until the model's intermediate layer $l$. The temporally weighted distance of a sample $x$ with respect to the query point $x_{query}$ is given by:

$$
Weighted\,Distance(x) = w(t) \cdot d(x).
$$

Adding this weighting function creates an additional computational cost of $\mathcal{O}(N(1+N)) = \mathcal{O}(N^2)$ per query to search the time indices for the query and entire data samples, which amounts to approximately 129.90 seconds per query on average on our machine specifications.

\begin{figure}[H]
    \centering
    \includegraphics[width=\linewidth]{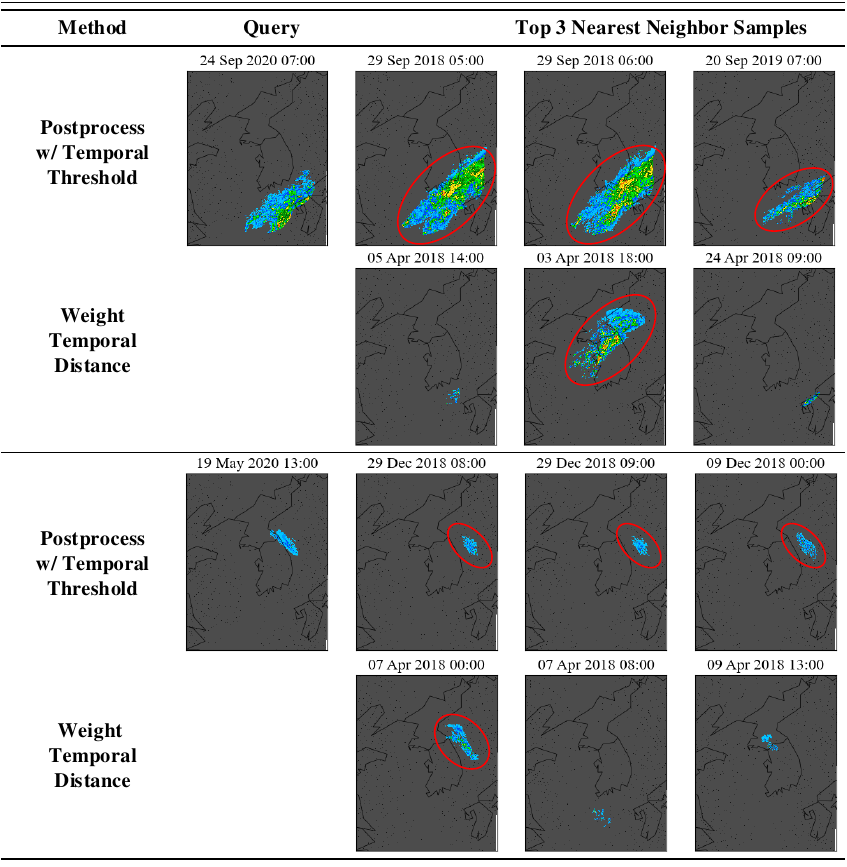}
    \caption{Nearest neighbor results from two query samples based on temporal distance weights and postprocessing with a temporal threshold of at least one month apart.}
    \label{fig:weighted_distance}
\end{figure}

As shown in Fig.~\ref{fig:weighted_distance}, the temporally weighted distance mitigates the issue of selecting temporally close samples. However, the nearest neighbor results degrades significantly. Combined with the additional computational cost, using temporally weighted distance does not seem viable. Instead, we add a post-processing procedure using a temporal distance threshold (e.g., at least one month apart).

\section{Wrapper Functions for Segmentation Models}\label{appx:wrapper}

Segmentation models generate pixel-wise class outputs, which is usually the same number of dimensions as the input.
To compute the importance scores $\Phi^{seg}: \mathbb{R}^{C_{\text{out}}\times W\times H} \mapsto \mathbb{R}^{C_{\text{in}}\times W\times H}$ \citep{simonyan2013deep} of a segmentation model $f^{seg}$, it is necessary to transform the multi-pixel outputs to scalar scores of each class by introducing a wrapper function $\Psi: \mathbb{R}^{C_{\text{out}}\times W\times H} \mapsto \mathbb{R}^{C_{\text{out}}}$. This transformation allows importance score to be computed as $\Phi^{seg} (f^{seg}, x) \Rightarrow \Phi(\Psi \circ f^{seg}, x).$
We introduce two generally used aggregation techniques in \cite{kokhlikyan2020captum} and design two additional techniques.

\subsection{Logit Sum.} 
This wrapper is introduced in \cite{kokhlikyan2020captum}.
It takes the sum of the entire logit values per output class channel. 
The logit value in a grid point means the model's confidence for the specific class, and the information of the confidence level of each pixel can be considered while summing output logits along the spatial axis since the model parameters are linked to being differentiable during backpropagation from the logit summed outputs to inputs. Mathematically:
\begin{linenomath*}
\begin{equation}
\Psi^c_{LogSum}(f, x) = \sum_{i}^{W} \sum_{j}^{H}  f(x_{i,j})_c
\end{equation}
\end{linenomath*}

One issue with this aggregation is that positive and negative logit values can cancel out one another, resulting in low logit values for target class. 

\subsection{Masked Sum.} 
This wrapper is also introduced in \cite{kokhlikyan2020captum}.
We address the limitations of summing raw logit values only for pixels that have been classified as target class.

\begin{linenomath*}
\begin{equation}
\Psi^k(f, x) =  \sum_{i}^{W} \sum_{j}^{H} f(x_{i,j})_k, 
\, \text{ such that } \operatornamewithlimits{argmax}_{k\in K} f(x_{i,j})_k =k 
\end{equation}
\end{linenomath*}

\subsection{Masked Scaled Sum.} 
The \textit{masked sum} technique can result in abnormally large importance score due to unnormalized large logit values.
We address this problem by scaling the logit sum by the predictive output mask size. In particular, since counting functions are not differentiable, we approximate it with the sum of applying Softmax operator to the logit value.
This wrapper works under the assumption that models trained with cross entropy objective tend to be overconfident, often resulting in Softmax value of almost 0 or 1.

\begin{linenomath*}
\begin{equation}
\begin{split}
\Psi^c_{ScaleSum}(f, x) =
& \frac{ \sum_{i}^{W} \sum_{j}^{H} f(x_{i,j})_c}{\|f(x_{i,j})_c\|} 
\approx \frac{ \sum_{i}^{W} \sum_{j}^{H} f(x_{i,j})_c}
{ \sum_{i}^{W} \sum_{j}^{H} \text{Softmax}f(x_{i,j})_c},  \\ 
& \text{ such that } \operatornamewithlimits{argmax}_{i,j} f(x_{i,j})=c
\end{split}
\end{equation}
\end{linenomath*}

\subsection{Masked Number of Pixels.} 
This technique only considers the number of pixels in the predictive mask, which means that it is equivalent to explaining how many pixels of a specific class has been predicted in the segmentation output, i.e., computing the denominator of the \textit{masked scaled sum} technique:
\begin{linenomath*}
\begin{equation}
\begin{split}
\Psi^c_{PixelNum}(f, x) 
& = \|f(x_{i,j})_c\| \approx { \sum_{i}^{W} \sum_{j}^{H} \text{Softmax}f(x_{i,j})_c},\\
& \text{ such that } \operatornamewithlimits{argmax}_{i,j} f(x_{i,j})=c 
\end{split}
\end{equation}
\end{linenomath*}

\appendix[E] 
\appendixtitle{Additional Results}
\section{Importance Score of Concept Activation Vectors}\label{appx:totalscores}

\begin{figure}[H]
    \centering
    \includegraphics[width=\linewidth]{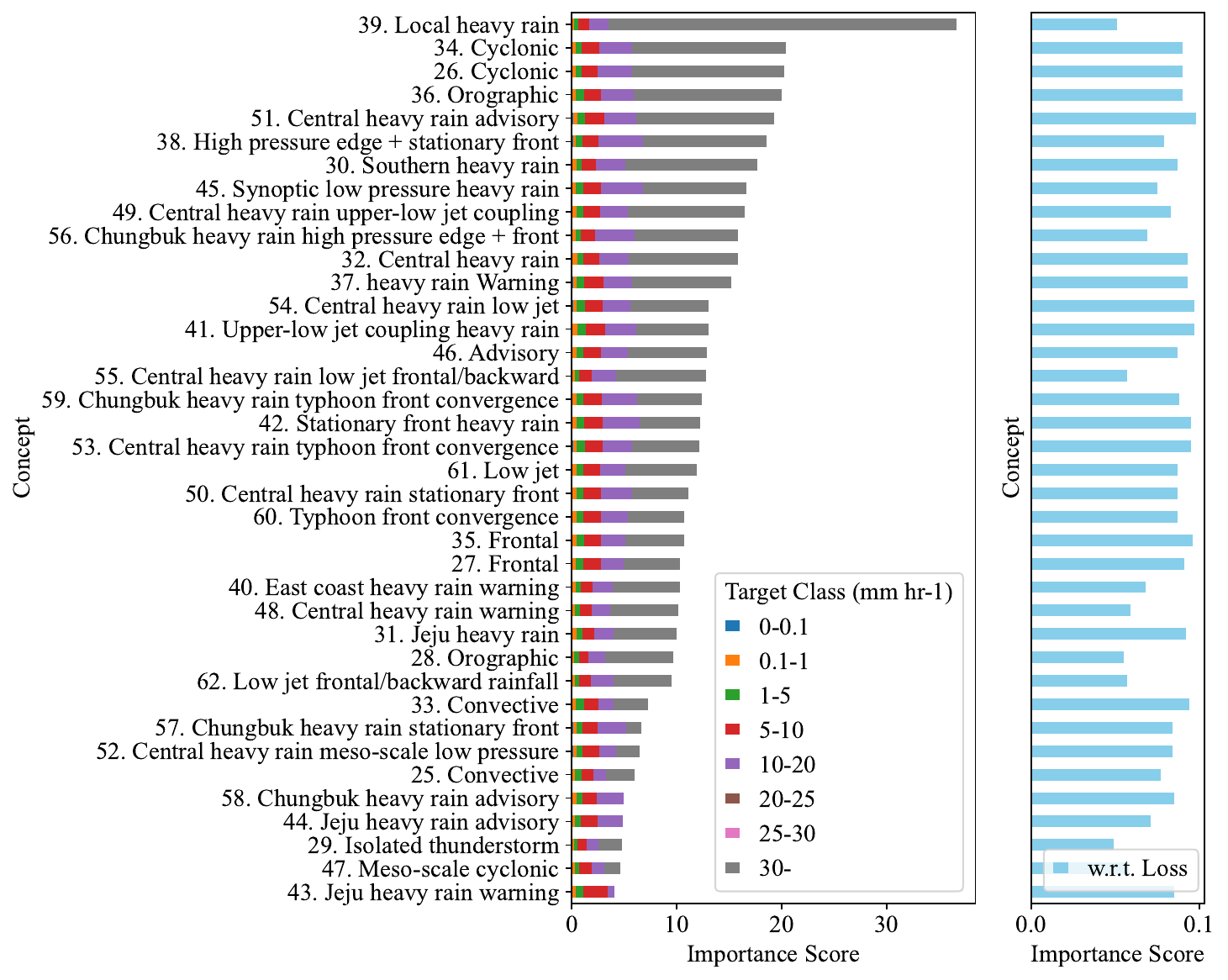}
    \caption{Importance score of concept activation vectors (continued from Fig.~\ref{fig:score}).} \label{fig:score2}
\end{figure}

The importance score of the remaining concepts are represented in Fig.~\ref{fig:score2}. 
The concept label indices of 0 to 24 in are annotated from the daily post-hoc forecast analysis reports. 
The concept label indices of 25 to 28 are from the Gaussian mixture model based rainfall classification results.
The concept label indices of 29 to 32 are from the self-organizing map model based rainfall classification results.
The concept label indices of 33 to 62 are from the heavy rainfall classification reports provided by NIMS.

\vspace{-0.2in}

%%%%%%%%%%%%%%%%%%%%%%%%%%%%%%%%%%%%%%%%%%%%%%%%%%%%%%%%%%%%%%%%%%%%%
% REFERENCES
%%%%%%%%%%%%%%%%%%%%%%%%%%%%%%%%%%%%%%%%%%%%%%%%%%%%%%%%%%%%%%%%%%%%%
% Make your BibTeX bibliography by using these commands:
% \bibliographystyle{ametsocV6}
% \bibliography{main}

\end{document}